\documentclass{article}

\usepackage{arxiv}

\usepackage[utf8]{inputenc} 
\usepackage[T1]{fontenc}    
\usepackage{hyperref}       
\usepackage{url}            
\usepackage{booktabs}       
\usepackage{amsfonts}       
\usepackage{nicefrac}       
\usepackage{microtype}      

\usepackage{amsmath}        
\usepackage{graphicx}       
\usepackage{amsthm}         
\usepackage{amssymb}        
\newcommand{\recdata}{\boldsymbol{\mathcal{D}}^r}
\usepackage[symbol, flushmargin]{footmisc}   
\usepackage{yhmath}         
\usepackage{multirow}       

\usepackage{comment}
\usepackage{color}
\usepackage{etoolbox}

\title{Towards Efficient and Secure Delivery of Data for Deep Learning with Privacy-Preserving}

\author{%
  Juncheng Shen\\
  College of Electrical Engineering\\
  Zhejiang University\\
  Hangzhou, Zhejiang, 310027, China\\
  \texttt{shenjc@zju.edu.cn} \\
  \And
  Juzheng Liu\\
  Department of Physics\\
  Tsinghua University\\
  Beijing, 100084, China\\
  \texttt{liu-jz15@mails.tsinghua.edu.cn}\\
  \And
  Yiran Chen and Hai Li\\
  Department of Electrical and Computer Engineering\\
  Duke University\\
  Durham, NC, 27708, USA\\
  \texttt{\{yiran.chen, hai.li\}@duke.edu}
}

\begin{document}

\maketitle

\begin{abstract}
Privacy recently emerges as a severe concern in deep learning, that is, sensitive data must be prohibited from being shared with the third party during deep neural network development.
In this paper, we propose \textit{Morphed Learning} (MoLe), an efficient and secure scheme to deliver deep learning data. 
MoLe has two main components: data morphing and Augmented Convolutional (Aug-Conv) layer. 
Data morphing allows data providers to send morphed data without privacy information, while Aug-Conv layer helps deep learning developers to apply their networks on the morphed data without performance penalty.
MoLe provides stronger security while introducing lower overhead compared to GAZELLE (USENIX Security 2018), which is another method with no performance penalty on the neural network. 
When using MoLe for VGG-16 network on CIFAR dataset, the computational overhead is only 9\% and the data transmission overhead is 5.12\%. 
As a comparison, GAZELLE has computational overhead of 10,000 times and data transmission overhead of 421,000 times.
In this setting, the attack success rate of adversary is $7.9 \times 10^{-90}$ for MoLe and $2.9 \times 10^{-30}$ for GAZELLE, respectively.
\end{abstract}

\section{Introduction}
\label{section_intro}
In recent years, deep learning has become the driving force for many artificial intelligence applications such as image classification~\cite{he2016deep}, object detection~\cite{he2017mask}, speed recognition~\cite{he2019streaming}, game intelligence~\cite{silver2018general}, etc. 
However, there are also many complaints about the substantially degraded performance of deep learning in real-world applications compared to that achieved in lab environments~\cite{marcus2018deep}. 
One common reason for the unsatisfactory performance of deep learning is the limited availability of high-quality training data, which is often caused by the privacy concern of the data. 
Take medical imaging as an example, 
researchers always struggle for obtaining health and medical data as these data contain private and sensitive information of patients; medical institutions are prohibited from sharing the data due to both legal and ethical concerns. 
The same difficulties happen in all the applications where privacy is involved, and greatly hinder adoptions of deep learning.

A general solution of the above challenge is to securely separate the privacy information from the data by paying extra computational and performance penalties. 
Several implementations based on \textit{Homomorphic Encryption} (HE)~\cite{gentry2009fully} or \textit{Secure Multi-Party Computation} (SMC)~\cite{yao1982protocols} have been proposed. 
However, these attempts suffer from the intrinsic limitation of their underlying cryptography techniques, and introduce computational and data transmission overheads that largely correlated with the depth of the target neural networks.
These overheads increase quickly with the depth of deep neural networks~\cite{he2017channel, sercu2016very} and make these privacy-preserving schemes impractical in some scenarios such as training. 
~\cite{osia2017hybrid, wang2018not} exploits property of the neural network itself and transmits low-level extracted features instead of the original data.
Although the feature transmitting methods can quarantine the original data, network performance is usually compromised.

In this paper, we propose \textit{Morphed Learning} (MoLe), a privacy-preserving scheme for efficient and secure delivery of deep learning data that can be used in both training and inference stages. The main contributions of our work are follows:


\begin{itemize}

\item
We propose \textit{Data Morphing} that performs linear shift of the original data in a multidimensional space. As the first key component of MoLe, Data Morphing blends each element of the data matrix with other elements in order to effectively hide sensitive information and prevent the morphed data from being recognized by human observers. 

\item 
We propose \textit{Augmented Convolutional (Aug-Conv) layer} that is composed of inverse transformation of data morphing and the first convolutional layer of the original neural network. As the second key component of MoLe, Aug-Conv layer can restore the performance degradation of the neural network incurred by the input shift in data morphing.



\item 
We verify the effectiveness of Aug-Conv layer by experiments. For CIFAR dataset~\cite{CIFAR10}, for instance, the penalty of training performance for a VGG-16 network is within the margin of error when replacing the first convolutional layer by Aug-Conv layer.

\item 
Our theoretical security analysis on three attack schemes against MoLe show that the upper bound of adversary's success probability is low even with a strict privacy reservation requirement. 
Specifically, the attack success probability for VGG-16 network is $P = 7.9 \times 10^{-90}$.

\item 
Our theoretical analysis proved that the computational and transmission overheads of MoLe are irrelevant to the depth of the neural network or the size of the dataset. 
Specifically, applying MoLe on a VGG-16 network for CIFAR dataset, the computational overhead and transmission overhead are 9\% and 5.12\% respectively.



\end{itemize}

The code to construct data morphing, Aug-Conv layer and to reproduce our experiments can be found at: \href{https://github.com/NIPS2019-authors/MoLe_public}{\url{https://github.com/NIPS2019-authors/MoLe_public}}

\section{Related work}

\textbf{HE based data delivery schemes.} 
The idea of HE is to compute on ciphertext and generate the ciphertext of result which can then be decrypted to original result. 
The latest work~\cite{sanyal2018tapas} reported that performing inference on encrypted MNIST dataset on a 16-core workstation took 2 hours, while performing inference on plain MNIST dataset on a regular PC takes only a few minutes.
Because of their high computational costs, 
the HE based schemes~\cite{gilad2016cryptonets, bourse2018fast, sanyal2018tapas} are applicable to only inference stage of deep learning, leaving the privacy information in training dataset unprotected.
HE based schemes also introduce constraints to the design of neural networks because both~\cite{bourse2018fast} and~\cite{sanyal2018tapas} require the range of activation to be discretized or even binarized.

\textbf{SMC based data delivery schemes.} 
In SMC, multiple parties hold their own private data, and wish to jointly compute a function which is dependent on data of all parties without each party sending its data to other parities.
Following the idea of SMC, distributed selective SGD was proposed~\cite{shokri2015privacy} and used a centralized parameter server for gradient transferring.
However, recent study~\cite{aono2018privacy} discovered a security flaw in~\cite{shokri2015privacy} as the parameter server is capable of recovering the training data from the gradients.
In addition, approaches combining HE and two-party computation (2PC), a subfield of SMC, were proposed to provide privacy-preserving for both training and inference stages~\cite{liu2017oblivious, rouhani2018deepsecure, juvekar2018gazelle}. 
However, the computational overhead for this kind of solutions is still high: the latest work~\cite{juvekar2018gazelle} reported $421,000\times$ data transmission overhead and more than $10,000\times$ execution time overhead, compared to non privacy-preserving method.

\textbf{Feature transmission based data delivery schemes.}
Unlike cryptography based schemes, feature transmission based schemes~\cite{osia2017hybrid, wang2018not} utilize the property of neural network itself.
Instead of sending the original data, the data provider computes several layers of the neural network and then send the extracted features.
This process reduces the practicality as the data provider needs to have computing capability to compute the network layers. 
Furthermore, features usually have more channels than the original data, which increase data transmission overhead.
To combat reverse engineering, noises are applied to the features, resulting in the degradation of network performance:~\cite{wang2018not} reported 62.8\% higher error rate for CIFAR-10 dataset.

\textbf{Differential privacy.} 
Differential privacy is a popular technique of privacy-preserving for deep learning applications, which usually serves the following two purposes: 
1. Preventing adversary from identifying or recovering training data from a published network model~\cite{phan2016differential};
2. Enhancing security of collaborative training schemes against reverse engineering on shared gradients~\cite{abadi2016deep}.
The first purpose is for defending inversion attacks, which is not in the scope of this paper;
The second purpose suffers from a fatal security vulnerably as malicious participant can breach the differential privacy using prototypical data generated by GANs~\cite{hitaj2017deep}.
For the above two reasons, we do not compare our work with differential privacy based privacy-preserving schemes.

\section{Preliminary}
\label{section_terms_and_settings}
\label{section_notations}
\textbf{Terms and settings}.
We imagine a scenario in which entity A owns a database of sensitive user information and entity B has a team of competent developers for deep learning applications.
We use the term \textit{data provider} to denote entity A and the term \textit{developer} to denote entity B.
In this scenario, the data provider wants to outsource the development task to the developer, and therefore he needs to transmit the data to the developer.
To minimize the risk of potential user privacy leakage, the data provider needs a method to hide the private information within the data.
On the other hand, the developer wishes the performance penalty or the number of network structural constraints incurred by the method to be as minimum as possible.

The developer can be also a potential adversary who can be benefited from recovering the private information hidden in the data. 
Since the data provider requires help from the developer to develop the deep learning applications, we can safely assume that the data provide does not (need to) have adequate resource for large-scale neural network training.

\textbf{Notations}.
Our notations follow two rules:
1. Bold font letters indicate matrices such as $\textbf{A}$, and $A_{x, y}$ refers to the element with coordinate of $(x, y)$ in the $\textbf{A}$.
The element at the top-left corner is considered to be the origin. 
$x$ axis indexes for rows and $y$ axis indexes for columns.
2. $\textbf{K}_{i, j}$ indicates a kernel for a convolutional layer, where $i$ is the index for input channels and $j$ is the index for output channels.
Combining the two rules, $k_{(i, j),(x, y)}$ stands for the element with coordinate $(x, y)$ in the $\textbf{K}_{i, j}$.
All the coordinates of matrices in this paper use zero based indexing.

\section{Method}

\begin{figure}[b]
\centering
\vspace{-12pt}
\includegraphics[width=5.5in]{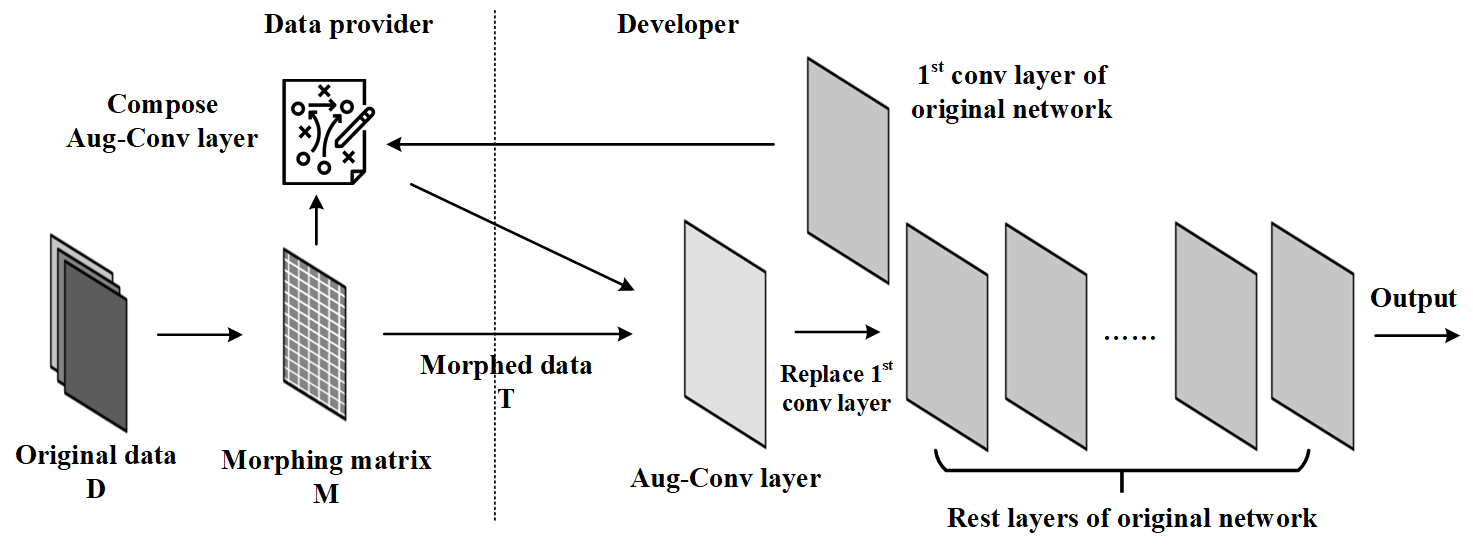}
\vspace{-12pt}
\caption{The process of utilizing MoLe for privacy-preserving.}
\label{fig_mole}
\vspace{-8pt}
\end{figure}

MoLe consists of two main components which are named as \textit{data morphing} and \textit{Aug-Conv layer}, respectively.
Data morphing serves the data provider by providing human unrecognizable data transformation. 
The computation of data morphing can be easily supported by regular CPUs.
Aug-Conv layer fully compensates the performance degradation of the neural network due to data morphing, and it is compatible to all convolutional neural network structures.
Figure~\ref{fig_mole} depicts the process of utilizing MoLe for privacy-preserving.
First, the developer trains his/her network on a public dataset similar to the dataset of data provider's, and sends the first trained convolutional layer to the data provider.
Second, the data provider generates morphing matrix and Aug-Conv layer, and uses data morphing to transform his/her data;
The developer replaces the first convolutional layer in his/her network, and then conducts training and inference on the morphed data without modifying any other parts of the network.
During the training process, the developer treats the Aug-Conv layer as a fixed feature extractor similarly to pre-trained layers in transfer learning~\cite{sharif2014cnn, yosinski2014transferable}.
In the rest of this paper, we assume the following attributions of the first convolution layer: the shape of input data is $m \times m$ with $\alpha$ channels, the shape of output features is $n \times n$ with $\beta$ channels and the shape of convolutional kernel is $p \times p$.

\subsection{Data to row vector}
Converting the first convolutional layer to an equivalent matrix multiplication is the basis of both data morphing and Aug-Conv layer.
Image to column matrix (im2col) is a trick to increase data parallelism for convolutional layer, and it has been widely implemented in various deep learning frameworks~\cite{jia2014caffe, abadi2016tensorflow, paszke2017automatic}.
The goal of im2col is to convert convolutional layer from sliding convolutional kernels on the input data to matrix-matrix multiplication~\cite{chellapilla2006high}.
d2r is a more extreme version of im2col that converts the convolutional layer to the product of a row vector and a large matrix. d2r includes the following steps:

1. Unroll the input data $\textbf{D}$ to row vectors. 
We first unroll the data of each channel by placing the row vector with a smaller row index on the left, and then concatenate each channel vector by placing the channel with a smaller channel index on the left.
The dimension of unrolled data vector $\textbf{D}^r$ is $1 \times \alpha m^2$. 
A figure showing the unrolling process can be found in our supplementary material.

2. Replace the convolution operation using a large matrix $\textbf{C}$ the dimension of which is $\alpha m^2 \times \beta n^2$.
We first initialize $\textbf{C}$ with zero elements.
Then, for every weight $k_{(i, j), (a, b)}$, we assign $C_{x, y} = k_{(i, j), (a, b)}$ when $(x, y)$ satisfies: 
\begin{equation}
\begin{split}
x &= im^2 + am + cm + b + d \\
y &= jn^2 + cn + d.
\end{split}
\label{eq_d2r_coordinate}
\end{equation}
Here $i \in [0, \alpha)$ is the index of input channels; $j \in [0, \beta)$ is the index of output channels; $(c, d)$ is the coordinate of input data and $c, d \in [0, m - p + 1)$; $a, b \in [0, p)$ are the coordinate of kernels.

3. After multiplying $\textbf{D}^r$ and $\textbf{C}$, we can get a row vector $\textbf{F}^r$ with a dimension of $1 \times \beta n^2$.
Following the reverse process in step 1, the features that is identical to the output of the original convolutional operation can be reconstructed.

The process of using d2r to perform the convolutional layer operation can be found in supplementary material. In the rest of the paper, a matrix with superscript $r$ represents its d2r unrolled row vector.

\subsection{Data morphing}
\label{section_data_morphing}
Data morphing is designed to meet the following three requirements.
1) \textit{Equal-sized input and output data}: This requirement ensures that the data transmission overhead does not have a correlation with the amount of data;
2) \textit{Adjustable computational cost}: This requirement comes from the setting in section~\ref{section_terms_and_settings}, as the data provider may have limited computing resources;
3) \textit{Unrecognizable transformation}: The output of data morphing should not be recognized by human observers for privacy-preserving purpose.

Before data morphing, the original data $\textbf{D}$ is unrolled to $\textbf{D}^r$ using d2r.
The data morphing operation can be described as:
\begin{equation}
\textbf{D}^r \cdot \textbf{M} = \textbf{T}^r.
\label{eq_data_morphing}
\end{equation}
Here $\textbf{T}^r$ is a row vector representing the morphed data. 
$\textbf{M}$ is the morphing matrix, which is constructed as follows:

1. Construct core morphing matrix $\textbf{M}'$. 
$\textbf{M}'$ is an invertible matrix, and all of its elements are random and non-zero.
Assume the dimension of $\textbf{M}'$ is $q \times q$, $q$ should satisfy: 
\begin{equation}
\kappa = \frac{\alpha m^2}{q} \in \mathbb{Z}+.
\label{eq_matrix_shape_requirement}
\end{equation}
Here $\kappa$ is called the morphing scale factor.

2. Construct $\textbf{M}$ by diagonally scaling $\textbf{M}'$ to $\alpha m^2 \times \alpha m^2$.
The relation between the two matrices can be written as:
\begin{equation}
M_{x,y} = 
\begin{cases}
M'_{x-Nq,y-Nq}, x \in [Nq, (N+1)q) \land y \in [Nq, (N+1)q)\\
0, \text{else}.
\end{cases}
\label{eq_m_scaling}
\end{equation}
Here $N$ satisfies $N \in \mathbb{N} \land N < \kappa$. 
A figure showing the process of diagonally scaling can be found in our supplementary materials.

The morphing scaling factor $\kappa$ controls the tradeoff between the computational cost and the effectiveness of privacy-preserving, as shown in our supplementary materials.
Since recent study~\cite{hassan2016performance} showed that modern desktop CPUs with support of AVX instructions can compute simple matrix multiplications with decent efficiency, it is recommended for desktop PC user to use a larger morphing core for better privacy. 
For lower-end compute devices such as embedded systems, however, data morphing is also able to adjust the required computing power by paying the cost of privacy. 

It is worth noting that the privacy-preserving feature offered by data morphing also requires secure storage of $\textbf{M}$. 
If the developer is able to acquire $\textbf{M}$, he could recover the original data by $\textbf{D}^r =\textbf{T}^r \cdot \textbf{M}^{-1}$, where $\textbf{M}^{-1}$ is the inverse matrix of $\textbf{M}$.

\subsection{Augmented convolutional (Aug-Conv) layer}
The design of Aug-Conv layer also needs to meet three requirements:
1) \textit{Equivalent feature extraction from the morphed data}: It ensures that the network trained on the morphed data can achieve the same performance as that trained on the original data. No modification on the network is needed;
2) \textit{Resistance to reverse engineering attack}: The Aug-Conv layer should not reveal the details of the morphing matrix $\textbf{M}$ under reverse engineering attack.
3) \textit{Resistance to reverse convolutional operation attack}: Although the Aug-Conv layer needs to extract equivalent features, it should not output identical features. Otherwise, the developer could recover the original data using the reverse operation of the convolutional layer.

\textbf{Inverse matrix multiplication}.
During d2r, the convolutional operation of the original first layer is converted to the matrix product $\textbf{F}^r = \textbf{D}^r \cdot \textbf{C}$.
The matrix representation of the convolutional operation allows us to form the Aug-Conv layer by matrix multiplication.
The detailed steps are:

1. Calculate inverse morphing matrix $\textbf{M}^{-1}$.
When using d2r for convolutional operation, $\textbf{M}^{-1}$ can restore the row vector of morphed data $\textbf{T}^r$ back to $\textbf{D}^r$, which can be represented as $\textbf{T}^r \cdot \textbf{M}^{-1} = \textbf{D}^r$.

2. Multiplying $\textbf{M}^{-1}$ and $\textbf{C}$ to form the Aug-Conv layer $\textbf{C}^{ac}$ as $\textbf{C}^{ac} = \textbf{M}^{-1} \cdot \textbf{C}$.

At this point, the features extracted by Aug-Conv layer from morphed data is identical to the features extracted by the convolutional layer from original data, or:
\begin{equation}
\textbf{T}^{r} \cdot \textbf{C}^{ac} = \textbf{T}^{r} \cdot \textbf{M}^{-1} \cdot \textbf{C} = \textbf{D}^{r} \cdot \textbf{C} = \textbf{F}^{r}.
\label{eq_aug_conv_extraction}
\end{equation}

The convolutional layer $\textbf{C}$ is trained on another similar dataset, which works essentially as a low-level feature extractor on that dataset. 
It can be reused for the dataset provided by the data provider, since low-level feature extractor for similar tasks usually shows strong transferability~\cite{yosinski2014transferable} and therefore, meets the the first requirement of Aug-Conv layer.
Inverse matrix multiplication also meets the above second requirement, as it hides the detailed value of each element in the inverse matrix $\textbf{M}^{-1}$.

\textbf{Channel order randomization}.
The $\textbf{C}^{ac}$ constructed from inverse matrix multiplication suffers the vulnerability of inverse convolutional operation attack.
As shown in Figure~\ref{fig_mole}, $\textbf{C}$ is provided by the developer and thus, he/she can also obtain its inverse $\textbf{C}^{-1}$.
As a result, the developer can recover the original data by: $\textbf{D}^{r} = \textbf{F}^{r} \cdot \textbf{C}^{-1}$.
To meet the third requirement of Aug-Conv layer, we randomly shuffle the order of output feature channels.
Assume $\textbf{F}' = rand(\textbf{F})$ is the output features with randomized channel order and $rand$ is the randomization function, $\textbf{F}'$ and $\textbf{F}$ are two sets of equivalent features for the rest layers of the neural network.
Here the shuffled channel order can be learned by the rest layers in training.
On the other hand, function $rand$ prevents the developer from restoring the original data as $\textbf{D}^{r} \neq rand(\textbf{F}^{r}) \cdot \textbf{C}^{-1}$, and therefore the third requirement is met.
Similarly to $\textbf{M}$, the detailed channel order used for $rand$ needs to be stored securely.

The following two steps reduce the $\textbf{C}^{ac}$'s vulnerability by implementing $rand$ function in the Aug-Conv layer:
1) Divide $\textbf{C}^{ac}$ into $\beta$ groups and each group contains $n^2$ continuous columns;
2) Randomly shuffle the order of the column groups to construct Aug-Conv layer matrix with feature randomization.

\subsection{Verification of the effectiveness of Aug-Conv layer}
During the construction of Aug-Conv layer, we assume that the channel order randomization of features can be learned in training process and would not harm the performance of the neural network.
However, it is unclear whether this assumption is true.
Therefore we use experiments on CIFAR dataset to verify the effectiveness of Aug-Conv layer.
We trained the networks with the same hyperparameter settings: learning rate is 0.001, total training epochs is 300, dropout probability is 0.5, batch size is 64, $\kappa = 1$ and learning rate decays by 0.5 for every 60 epochs.
The results are summarized in Table~\ref{tab_result}.
The experimental results verify that Aug-Conv layer can help a network to achieve the performance on morphed data similar to its original network on clean data.
On the contrary, without the help of Aug-Conv layer, network performance would drop significantly on morphed data.

\begin{table}[!t]
\vspace{-8pt}
\caption{Aug-Conv layer effectiveness experiments}
\vspace{-4pt}
\label{tab_result}
\centering
\begin{tabular}{|c|c|c|c|}
\hline
Dataset                    & Data morphing & Network & Accuracy \\ \hline
\multirow{3}{*}{CIFAR-10}  & No            & VGG-16 original             & 87.2\%   \\ \cline{2-4} 
                           & Yes           & VGG-16 with Aug-Conv layer            & 87.4\%   \\ \cline{2-4} 
                           & Yes           & VGG-16 original             & 15.8\%   \\ \hline
\multirow{3}{*}{CIFAR-100} & No            & VGG-16 original             & 55.9\%   \\ \cline{2-4} 
                           & Yes           & VGG-16 with Aug-Conv             & 56.4\%   \\ \cline{2-4} 
                           & Yes           & VGG-16 original             & 2.01\%   \\ \hline
\end{tabular}
\vspace{-16pt}
\end{table}
\section{Theoretical analysis on security and overhead}
\label{section_analysis}
\subsection{Threat model}
In this subsection, two threat models are introduced for security analysis.
The developer is considered to be a potential \textit{Honest-but-Curious} (HBC) adversary.
HBC is a typical threat model for SMC where the adversary tries to discover as much information from other parties as possible without violating the protocol.
The HBC setting for MoLe means that the developer tries to recover the original data using the information sent from the data provider, including $\textbf{T}$ and $\textbf{C}^{ac}$.
The developer does not exploit other security breaches to access $\textbf{M}^{-1}$, $rand$ or $\textbf{D}$.

Additionally, \textit{Semi-Honest-but-Curious} (SHBC) adversarial model is another stronger threat model in which the developer is capable of acquiring several $\textbf{D}$-$\textbf{T}$ pairs.
The SHBC model allows further analysis on the situations where the adversary managed to inject some of his/her own data into the data provider's database beforehand.

\subsection{Security analysis}
In this subsection, the security of MoLe against three possible attack methods is quantitatively analyzed in the form of attack success probability.
The three attack methods are \textit{Brute Force Attack}, \textit{Aug-Conv Reversing Attack} and \textit{\textbf{D}-\textbf{T} Pair Attack}.


\theoremstyle{definition}

\newtheorem{definition}{Definition}
\begin{definition}
Unit $l^2$-norm matrix. Any matrix $\textbf{A}$ that satisfies $|\textbf{A}|_2 = 1$ is a unit $l^2$-norm matrix, where $|\textbf{A}|_2$ denotes the $l^2$-norm of $\textbf{A}$.
\end{definition}

We first scale $\textbf{D}^r$ and each column in $\textbf{M}$ to be unit $l^2$ matrices in order to remove the measurement units of the convolutional layer.

\textbf{Brute Force Attack}.
The most straightforward attack to MoLe in HBC setting is to apply brute force attack on $\textbf{M}$.
Unlike encryption keys which would be useless even with one-bit difference, a close approximation of $\textbf{M}$ can also somehow recover recognizable data $\recdata$ as:
\begin{equation}
\label{eq_approximation}
\textbf{G} \approx \textbf{M} \Rightarrow 
\textbf{G}^{-1} \approx \textbf{M}^{-1} \Rightarrow
\recdata = \textbf{T}^r \cdot \textbf{G}^{-1} \approx \textbf{T}^r \cdot \textbf{M}^{-1} = \textbf{D}^r.
\end{equation}
Here $\textbf{G}$ is attacker's approximation of $\textbf{M}$.

\newtheorem{lemma}{Lemma}
\begin{lemma}
\label{lemma_1}
Assume $\textbf{A}$ is a given unit $l^2$-norm matrix which has $N$ elements, $P_1$ is the probability of a random unit $l^2$-norm $\textbf{B}$ that satisfies: 
\begin{equation}
\label{eq_l2_def}
|\textbf{A} - \textbf{B}|_2 \leq d \leq 1.
\end{equation}
The upper bound of $P_1$ can be calculated by:
\begin{equation}
\label{eq_p_upper_bound}
P_1 \leq \frac{1}{2}d^{(N - 1)}.
\end{equation}
\end{lemma}

\begin{lemma}
\label{lemma_2}
Assume $E_{rms}(\textbf{D}^r, \recdata)$ is the root mean square error between $\textbf{D}^r$ and $\recdata$, $N$ is the number of elements in $\textbf{M}$, $P_2$ is the probability of $E_{rms}(\textbf{D}^r, \recdata) \leq \frac{\sigma}{\sqrt[4]{N}}$.
$P_2$ satisfies:
\begin{equation}
P_2 = P_1(d = \sigma).
\end{equation}
Here $P_1$ is the probability in lemma~\ref{lemma_1}. The proof for lemma~\ref{lemma_1} and~\ref{lemma_2} can be found in our  supplementary material.
\end{lemma}

\newtheorem{theorem}{Theorem}

Substitute lemma~\ref{lemma_1} into lemma~\ref{lemma_2}, we can prove theorem~\ref{theorem_1} as:
\begin{theorem}
\label{theorem_1}
For original data $\textbf{D}^r$ and recovered data $\recdata$,  $\textbf{M}$ has $N = (\frac{\alpha m^2}{\kappa})^2$ elements, the upper bound of the probability $P_{M,bf}$ for $E_{rms}(\textbf{D}^r, \recdata) \leq \frac{\sigma}{\sqrt[4]{N}}$ is:
\begin{equation}
\label{eq_P_final}
P_{M,bf} \leq \frac{1}{2}\sigma ^{N-1} = \frac{1}{2}\sigma ^{(\frac{\alpha m^2}{\kappa})^2-1}.
\end{equation}
\end{theorem}

Here $\sigma$ is defined as the privacy reservation $R_p$ of MoLe, directing the maximum resemblance between $\textbf{D}^r$ and $\recdata$ allowed by the data provider.
$R_p \in (0, 1)$. A larger $R_p$ implies stricter requirement for privacy-preserving.

The adversary can also apply brute force attack on $rand$.
There are $\beta !$ possible randomization orders and therefore the success probability for brute force attack on $rand$ is $P_{r,bf} = \frac{1}{\beta !}$.

To give an intuitive understanding about the probabilities, we take CIFAR dataset as an example. 
Assume the data provider requires $R_p = 50\%$ and chooses $\kappa = 1$, and the network used by the data provider is VGG-16\cite{simonyan2014very}.
In this setting, $N = \frac{\alpha^2 m^4}{\kappa^2} = 3072^2$ and we can get $P_{M,bf} \leq 2^{-3072^2} \approx 2^{-9 \times 10^6}$, and $P_{r,bf} = (64!)^{-1} \approx 7.9 \times 10^{-90}$.
It is worth noting that $R_p = 50\%$ is already a very strict privacy-preserving requirement. 
The visualization result for $\textbf{D}^r$-$\recdata$ pairs with different $R_p$ can be found in our supplementary material.

\textbf{Aug-Conv Reversing Attack}.
Aug-Conv reversing attack is another possible attack in HBC setting. 
The adversary tries to factorize $\textbf{C}^{ac}$ to acquire $\textbf{M}^{-1}$ and $rand(\textbf{C})$.
Consider the channel with index of $j$ in the output features, the factorization process is to solve the equation set:
\begin{equation}
\label{eq_attack_factorization}
\textbf{C}^{ac}_{y}= \textbf{M}^{-1} \cdot \textbf{C}_{y}.
\end{equation}
Here $\textbf{C}^{ac}_{y}$ and $\textbf{C}_{y}$ denote the column vector with the index of $y$ in $\textbf{C}^{ac}$ and $\textbf{C}$, respectively, $y \in [jn^2, (j+1)n^2)$.
In equation set~(\ref{eq_attack_factorization}), $\textbf{M}^{-1}$ is unknown to the adversary, introducing $\frac{\alpha m^2}{\kappa}$ unknown vector variables.
In HBC settings, the adversary does not have access to $rand$ and therefore $rand(\textbf{C})$ is also unknown, introducing $\alpha \beta p^2$ unknown elemental variables.
The total number of unknown variables for the adversary is:
\begin{equation}
N_{unk} = \frac{\alpha m^2}{\kappa} + \alpha p^2 > \frac{\alpha m^2}{\kappa}.
\end{equation}
The number of equations in equation set~\ref{eq_attack_factorization} is $N_{eq} = n^2$.
To prevent the adversary from solving the equation set, $N_{unk}$ should be larger than $N_{eq}$, and therefore $\kappa$ should satisfy:
\begin{equation}
\frac{\alpha m^2}{\kappa} \geq n^2 \Rightarrow \kappa \leq \frac{\alpha m^2}{n^2}.
\label{eq:k_upperbound}
\end{equation}
Equation~(\ref{eq:k_upperbound}) gives the upper bound of $\kappa$.
Additionally, even if the adversary includes more input channels into the equation set, each channel can only provide $\alpha p^2$ independent equations while also introducing another $\alpha p^2$ unknown variables.
Thus, including more input channels does not help the adversary to solve the equation set.

The adversary can use equation set~(\ref{eq_attack_factorization}) to represent $N_{eq}$ elements in a column vector of $\textbf{M}^{-1}$ by other elements, and thus eliminating $n^2$ independent elements in each column.
The adversary can then use brute force attack to $\textbf{M}^{-1}$ with reduced independent elements to achieve a higher success probability $P_{M, ar}$, which can be represented as:
\begin{equation}
\label{eq_p_m_ar}
P_{M, ar} = P_{M, bf}\Big(N = (\frac{\alpha m^2}{\kappa} - n^2)(\frac{\alpha m^2}{\kappa}) + \alpha p^2\Big) \leq \frac{1}{2}\sigma ^{(\frac{\alpha m^2}{\kappa}-n^2)(\frac{\alpha m^2}{\kappa})+\alpha p^2-1}.
\end{equation}
With the same CIFAR dataset and VGG-16 network setting adopted in the analysis of brute force attack, we can get $P_{M,ar} \leq 2^{-3072 \times 2048} \approx 2^{-6 \times 10^6}$.
For more complicated tasks like ImageNet~\cite{deng2009imagenet}, $P_{M,ar} \leq 2^{-7.5 \times 10^9}$ which is negligible due to the increase of $m$ and $n$.

Combining $P_{M,bf}$, $P_{r,bf}$ and $P_{M,ar}$, we can see that for most datasets and networks, $P_{r,bf}$ is the upper boundary of attacker's success probability in HBC setting. 

\textbf{D-T Pair Attack}.
$\textbf{D}$-$\textbf{T}$ pair attack applies only to the SHBC threat model.
In this attack, we use the number of $\textbf{D}^r$-$\textbf{T}^r$ pairs required by the adversary to evaluate the security.
In $\textbf{D}$-$\textbf{T}$ pair attack, the adversary aims to calculate $\textbf{M}'$ by: 
\begin{equation}
\label{eq_d-t_pair_attack}
\textbf{M}' = \mathbb{D}^{-1} \cdot \mathbb{T}.
\end{equation}
Here $\mathbb{D}$ and $\mathbb{T}$ are constructed by stacking multiple $\textbf{D}^r$-$\textbf{T}^r$ pairs correspondingly.
We can see from equation~(\ref{eq_d-t_pair_attack}) that the required number of $\textbf{D}$-$\textbf{T}$ pair equals the umber of rows in $\textbf{M}'$, which is $q = \frac{\alpha m^2}{\kappa}$.
In the same setting of brute force attack, the attack requires 3,072 $\textbf{D}^r$-$\textbf{T}^r$ pairs.

\subsection{Overhead analysis}
\textbf{Computational overhead}.
The computational overhead introduced by MoLe can be divided into two parts.
The first part is on the data provider's side and caused by the calculation of equation~(\ref{eq_data_morphing}).
The second part is on the developer's side and incurred by replacing $\textbf{C}$ with $\textbf{C}^{ac}$.
We use the number of extra \textit{Multiply–Accumulate} (MAC) operations to measure the computational overhead.
For the data provider, the total number of MAC operations $O_{comp, dp}$ that required by each calculation of Equation~(\ref{eq_data_morphing}) can be represented by: 
\begin{equation}
\label{eq_O_mac_dp}
O_{comp, dp} = \alpha q^2.
\end{equation}
Here multiplications with zero element are omitted.
For the developer, the computational overhead $O_{comp, dev}$ is the difference between the MAC operations needed by calculating $\textbf{T}^r \cdot \textbf{C}^{ac}$ and $\textbf{D}^r \cdot \textbf{C}$, which can be calculated by: 
\begin{equation}
\label{eq_O_mac_dev}
O_{comp, dev} = \alpha m^2 \beta n^2 - \alpha p^2 \beta n^2 = (m^2 - p^2)\alpha \beta n^2.
\end{equation}
We can see from Equation~(\ref{eq_O_mac_dp}) and~(\ref{eq_O_mac_dev}) that neither $O_{comp, dp}$ nor $O_{comp, dev}$ is related to the depth of the neural network.
On the other hand, we can also see that the computational overhead grows quadratically with the size of input data $m$ and output feature $n$, making MoLe more suitable for low dimensional tasks.
To put it in perspective, the computational overhead proportion to the original network is 9\% for VGG-16 network on CIFAR dataset, and 10$\times$ for ResNet-152 network on ImageNet dataset.

\textbf{Data transmission overhead}.
Since $\textbf{D}^r$ and $\textbf{T}^r$ has the same dimension, the data transmission overhead $O_{data}$ is determined by the size of $\textbf{C}^{ac}$, i.e.,
$O_{data} = (\alpha m^2)^2$ or the number of elements in $\textbf{C}^{ac}$.
$O_{data}$ is irrelevant to either the size of the dataset or the depth of the neural network.
For CIFAR-10 and CIFAR-100, $O_{data}$ is 5.12\% to the whole dataset. 
For a larger dataset like ImageNet, $O_{data}$ is merely about 1\%.
The performance penalty and overhead comparison of MoLe and other related methods on VGG-16 network and CIFAR-10 dataset is summarized in Table~\ref{tab_1}.

\begin{table}[!tb]
\centering
\renewcommand\arraystretch{1}
\vspace{-4pt}
\caption{MoLe versus other related methods (CIFAR-10 dataset and VGG-16 network)}
\vspace{-4pt}
\label{tab_1}
\begin{tabular}{|c|c|c|c|c|}
\hline
Method                                                                 &\begin{tabular}[c]{@{}c@{}}Performance \\ penalty\end{tabular}&\begin{tabular}[c]{@{}c@{}}Data transmission \\ overhead\end{tabular}&\begin{tabular}[c]{@{}c@{}}Computational \\ overhead\end{tabular}& \begin{tabular}[c]{@{}c@{}}Attack success \\ probability\end{tabular}\\ \hline
MoLe                                                                   & 0                                                                   & 0.05$\times$                          & 0.09$\times$& $7.9 \times 10^{-90}$                    \\ \hline
SMC based\cite{juvekar2018gazelle}                                                              & 0                                                                   & 421,000$\times$                    & 10,000$\times$&$2.9 \times 10^{-39}$                 \\ \hline
\begin{tabular}[c]{@{}c@{}}Feature trans-\\ mission based\cite{wang2018not}\end{tabular} & \begin{tabular}[c]{@{}c@{}}0.628$\times$ higher \\ error rate\end{tabular} & 64$\times$                         & 0 & Not reported                     \\ \hline
\end{tabular}
\vspace{-16pt}
\end{table}

\section{Conclusion}
In this work, we propose Mole -- an efficient and secure method for deep learning data delivery. 
MoLe includes two components: data morphing and Aug-Conv layer. 
Data morphing generates unrecognizable morphed data with low and adjustable computational cost, and Aug-Conv layer compensates the morphed data incurred performance degradation of neural networks.
Quantitative analysis on security and overhead shows that MoLe offers a very strong security protection with low overheads. 

\newpage
\bibliographystyle{unsrt}
\bibliography{reference.bib}

\newpage
\section*{APPENDIX A. PROOFS}
\subsection*{Proof for lemma 1}
\begin{figure}[!h]
\centering
\includegraphics{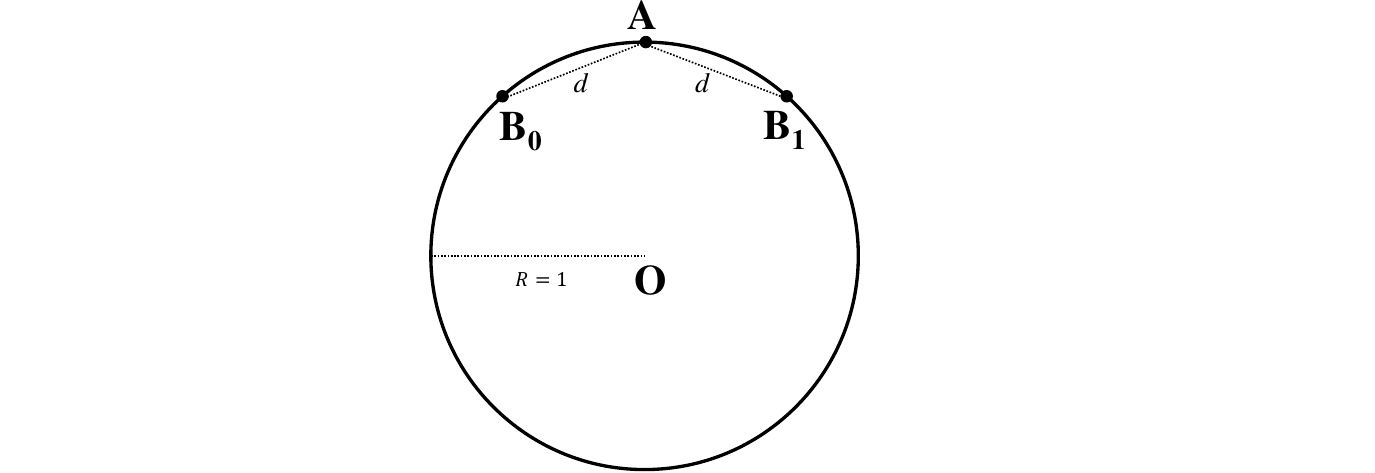}
\vspace{-12pt}
\caption{2-D projection of the $N$-D space}
\label{fig_2-D_projection}
\vspace{-6pt}
\end{figure}

Both $\textbf{A}$ and $\textbf{B}$ can be seen as points in an $N$-dimensional ($N$-D) space.
Since both of them are unit $l^2$-norm matrices, they land on the surface of an $N$-D hypersphere.
Figure~\ref{fig_2-D_projection} shows the 2-D projection of the $N$-D space.

$\textbf{B}_0$ and $\textbf{B}_1$ are the borders of $\textbf{B}$ which suffice $|\textbf{A} - \textbf{B}|_2 \leq d \leq 1$.
In the $N$-D space, assume the surface area of $\odot \textbf{O}$ is $S_{\textbf{O}}$, surface area of hypersurface $\wideparen{\textbf{B}_0\textbf{AB}_1}$ is $S_{\textbf{B}_0\textbf{AB}_1}$, then:
\begin{equation}
P_1 = \frac{S_{\textbf{B}_0\textbf{AB}_1}}{S_{\textbf{O}}}.
\label{eq_p_1}
\end{equation}
Now we make an auxiliary sphere $\odot \textbf{A}$ with $r_{\textbf{A}} = d$, as shown in figure~\ref{fig_auxiliary_sphere}:
\begin{figure}[!h]
\centering
\includegraphics{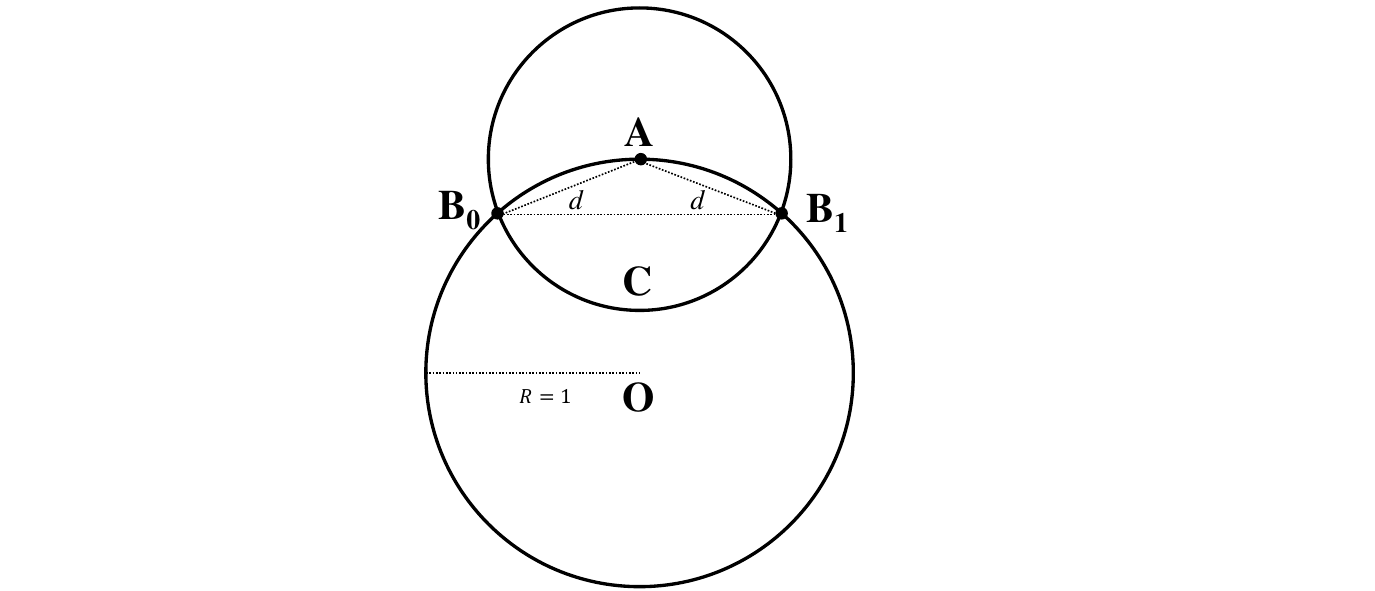}
\vspace{-12pt}
\caption{Add the auxiliary sphere}
\label{fig_auxiliary_sphere}
\vspace{-6pt}
\end{figure}

In the 2-D projection, we can see that the length of arc $\wideparen{\textbf{B}_0\textbf{AB}_1}$ is shorter than the length of arc $\wideparen{\textbf{B}_0\textbf{CB}_1}$, as the latter is further away from the straight line $\overline{\textbf{B}_0\textbf{B}_1}$.
Similarly, in the $N$-D space, the area of hypersurface $\wideparen{\textbf{B}_0\textbf{AB}_1}$ is smaller than the area of hypersurface $\wideparen{\textbf{B}_0\textbf{CB}_1}$.
Therefore, we have: 
\begin{equation}
P_1 \leq \frac{S_{\textbf{B}_0\textbf{CB}_1}}{S_{\textbf{O}}} \leq \frac{\frac{1}{2}S_{A}}{S_{\textbf{O}}}.
\label{eq_p_1_less_than_long}
\end{equation}
Here $S_{A}$ is the surface area of $\odot \textbf{A}$.
The $N$-D hypersphere surface area can be calculated as: 
\begin{equation}
S_N(R) = \frac{2 \pi ^{\frac{N+1}{2}}}{\Gamma(\frac{N+1}{2})}R^{N-1}.
\label{eq_hypersphere_area}
\end{equation}
Substitute Equation~(\ref{eq_hypersphere_area}) into Equation~(\ref{eq_p_1_less_than_long}), we can get:
\begin{equation}
P_1 \leq \frac{1}{2} d^{N-1}.
\end{equation}
Therefore, lemma 1 is proven.

\subsection*{Proof for lemma 2}
The sum of squared error (SSE) between $\textbf{D}^r$ and $\recdata$ can be calculated by: 
\begin{equation}
\label{eq_def_sse}
S(\textbf{D}^r, \recdata) = (\textbf{D}^r - \recdata)^T \cdot (\textbf{D}^r - \recdata).
\end{equation}
Substitute $\textbf{D}^r = \textbf{T}^r \cdot \textbf{M}^{-1}$ and $\recdata = \textbf{T}^r \cdot \textbf{G}$ into Equation~\ref{eq_def_sse} we can get:
\begin{equation}
\begin{split}
S(\textbf{D}^r, \recdata) &= (\textbf{T}^r \cdot \textbf{M}^{-1} - \textbf{T}^r \cdot \textbf{G})^{T} \cdot (\textbf{T}^r \cdot \textbf{M}^{-1} - \textbf{T}^r \cdot \textbf{G})\\
&= \big(\textbf{T}^r \cdot (\textbf{M}^{-1} - \textbf{G})\big)^{T} \cdot \big(\textbf{T}^r \cdot (\textbf{M}^{-1} - \textbf{G})\big)\\
&= \textbf{T}^{rT} \cdot (\textbf{M}^{-1} - \textbf{G})^{T} \cdot (\textbf{M}^{-1} - \textbf{G}) \cdot \textbf{T}^r.
\end{split}
\label{eq_sse_detail}
\end{equation}
For easy representation, assert $\textbf{K} = (\textbf{M}^{-1} - \textbf{G})^T \cdot (\textbf{M}^{-1} - \textbf{G})$. 
Equation~\ref{eq_sse_detail} can be simplified as: 
\begin{equation}
S(\textbf{D}^r, \recdata) = \textbf{T}^{rT} \cdot (\textbf{M}^{-1} - \textbf{G})^{T} \cdot (\textbf{M}^{-1} - \textbf{G}) \cdot \textbf{T}^r = \textbf{T}^{rT} \cdot \textbf{K} \cdot \textbf{T}^r.
\label{eq_sse_simple}
\end{equation}
Since $\textbf{K}$ is a symmetrical matrix, it can be factorized into the product of unitary matrix and diagonal matrix like: 
\begin{equation}
\textbf{K} = \textbf{P}^T \cdot \boldsymbol{\Lambda} \cdot \textbf{P}.
\label{eq_symmetric_matrix}
\end{equation}
Here $\textbf{P}$ is an unitary matrix and $\boldsymbol{\Lambda}$ is a diagonal matrix.
Substitute Equation~(\ref{eq_symmetric_matrix}) into Equation~(\ref{eq_sse_detail}), we have: 
\begin{equation}
\begin{split}
S(\textbf{D}^r, \recdata) &= \textbf{T}^{rT} \cdot \textbf{P}^{T} \cdot \boldsymbol{\Lambda} \cdot \textbf{P} \cdot \textbf{T}^r\\
&= \sum_{i=0}^{N'-1} \boldsymbol{\Lambda}_{ii}(\textbf{P} \cdot \textbf{T}^r)_i^2.
\end{split}
\label{eq_sse_final}
\end{equation}
Here $N'$ is the number of elements in $\textbf{T}^r$.
Since $\textbf{T}^r$ is the linear transformation of a data from real-world, we assume the elements in $\textbf{T}^r$ is approximately independent and identically distributed, then we can get:
\begin{equation}
\sum_{i=0}^{N'-1} E\big((\textbf{P} \cdot \textbf{T}^r)_i^2\big) = 1.
\label{eq_E_P_T}
\end{equation}
Here $E$ is the expectation.
Additionally, based on Equation~(\ref{eq_sse_final}), we can have: 
\begin{equation}
E\big(S(\textbf{D}^r, \recdata)\big) = \sum_{i=0}^{N'-1} \boldsymbol{\Lambda}_{ii} E\big((\textbf{P} \cdot \textbf{T}^r)_i^2\big).
\label{eq_E_sse}
\end{equation}
Combining equation~\ref{eq_E_P_T} and~\ref{eq_E_sse}, we can get:
\begin{equation}
\begin{split}
&\sum_{i=0}^{N'-1} E\big((\textbf{P} \cdot \textbf{T}^r)_i^2\big) = 1 \\
& \Rightarrow E\big((\textbf{P} \cdot \textbf{T}^r)_i^2\big) = \frac{1}{N'}\\
& \Rightarrow E\big(S(\textbf{D}^r, \recdata)\big) = \frac{\sum_{i=0}^{N'-1} \boldsymbol{\Lambda}_{ii}}{N'} = \frac{Tr(\boldsymbol{\Lambda})}{N'}.
\end{split}
\label{eq_sse_expectation}
\end{equation}
Here $Tr$ is the trace of matrix, i.e., the sum of all diagonal elements.
According to Equation~(\ref{eq_symmetric_matrix}), $Tr(\boldsymbol{\Lambda}) = Tr(\textbf{K})$.
Therefore we can represent $Tr(\boldsymbol{\Lambda})$ using $\textbf{G}$ and $\textbf{M}^{-1}$ as:
\begin{equation}
Tr(\boldsymbol{\Lambda}) = Tr(\textbf{K}) = \sum_{i=0}^{N''-1}(\textbf{M}^{-1}-\textbf{G})^T_i \cdot (\textbf{M}^{-1}-\textbf{G})_i.
\label{eq_trace_Lambda}
\end{equation}
Here $(\textbf{M}^{-1}-\textbf{G})_i$ denotes the column vector in $(\textbf{M}^{-1}-\textbf{G})$ with the index of $i$, $N''$ is the number of columns in $(\textbf{M}^{-1}-\textbf{G})$.
According to the rules of d2r, $N'' = N'$.
Substitute Equation~(\ref{eq_trace_Lambda}) into Equation~(\ref{eq_sse_expectation}), we have:
\begin{equation}
\begin{split}
E\big(S(\textbf{D}^r, \recdata)\big) &= \frac{1}{N'}\sum_{i=0}^{N'-1}(\textbf{M}^{-1}-\textbf{G})^T_i \cdot (\textbf{M}^{-1}-\textbf{G})_i\\
&= \frac{1}{N'}\sum_{i=0}^{N'-1}\sum_{j=0}^{N'-1}(\textbf{M}^{-1}-\textbf{G})_{i,j}^2.
\end{split}
\label{eq_e_sse_in_M_and_G}
\end{equation}
At this point, we have successfully established the relation between $S(\textbf{D}^r, \recdata)$ and $(\textbf{M}^{-1}-\textbf{G})$.
Additionally, we scale the norm of $\textbf{M}^{-1}$ and $\textbf{G}$ to $N'$ as:
\begin{equation}
\sum_{i=0}^{N'-1}\sum_{j=0}^{N'-1}(\textbf{M}^{-1}_{i,j})^2 =  \sum_{i=0}^{N'-1}\sum_{j=0}^{N'-1}\textbf{G}^2_{i,j} = N'.
\label{eq_normalize_M_and_G}
\end{equation}
The normalized $\textbf{M}^{-1}$ and $\textbf{G}$ can be considered as two points in a hypersphere with the radius of $R = \sqrt{N'}$.
The $l^2$ distance $d$ between $\textbf{M}^{-1}$ and $\textbf{G}$ is:
\begin{equation}
d = \sqrt{\sum_{i=0}^{N'-1}\sum_{j=0}^{N'-1}(\textbf{M}_{norm}^{-1}-\textbf{G}_{norm})_{i,j}^2}.
\label{eq_l2_distance}
\end{equation}
Combining Equation~(\ref{eq_e_sse_in_M_and_G}) and Equation~(\ref{eq_l2_distance}), we can get:
\begin{equation}
d = \sqrt{E\big(S(\textbf{D}^r, \recdata)\big)N'}.
\label{eq_d_equals}
\end{equation}
According to the relation between SSE and root mean square error, $E\big(S(\textbf{D}^r, \recdata)\big) = E_{rms}^2(\textbf{D}^r, \recdata)  N' \leq \sigma ^2 N'/{\sqrt{N}} $.
Since $N$ is the number of elements in $\textbf{M}$ and $N'$ is the number of elements in $\textbf{T}^r$, we have $\sqrt{N} = N'$, and therefore $E\big(S(\textbf{D}^r, \recdata)\big) \leq \sigma^2$.
Equation~(\ref{eq_d_equals}) can now be written as:
\begin{equation}
d \leq \sigma \sqrt{N'}.
\end{equation}
Re-scale the space by the factor of $\frac{1}{\sqrt{N'}}$, we can get $d \leq \sigma$ and $R = 1$.

Additionally, the dimension of the space equals the total number of elements in $\textbf{M}^{-1}$, which is $N$.
Therefore, $P_2 = P_1(d = \sigma)$ and lemma 2 is proven.

\newpage

\section*{APPENDIX B. ADDITIONAL FIGURES}
\begin{figure}[!h]
\centering
\includegraphics{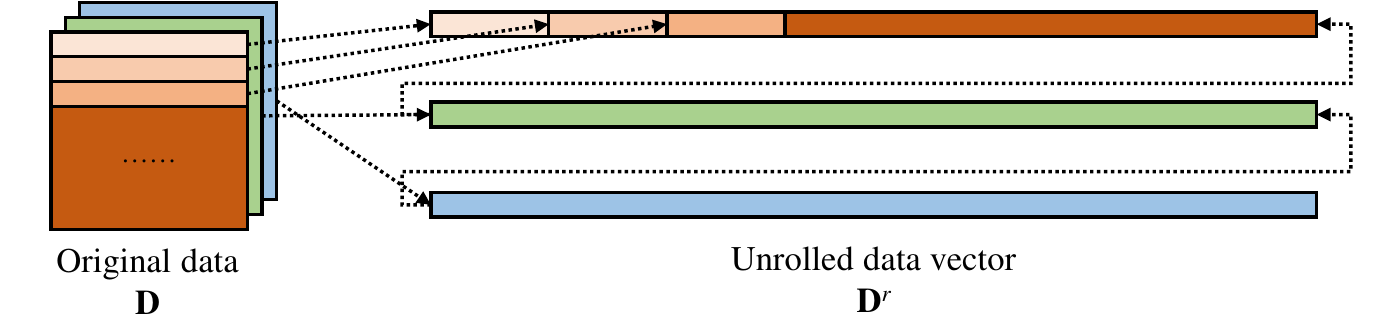}
\caption{Unrolling data for d2r.}
\label{fig_unroll_data}
\end{figure}

Figure~\ref{fig_unroll_data} shows the data unrolling process for d2r.

\begin{figure}[!h]
\centering
\includegraphics{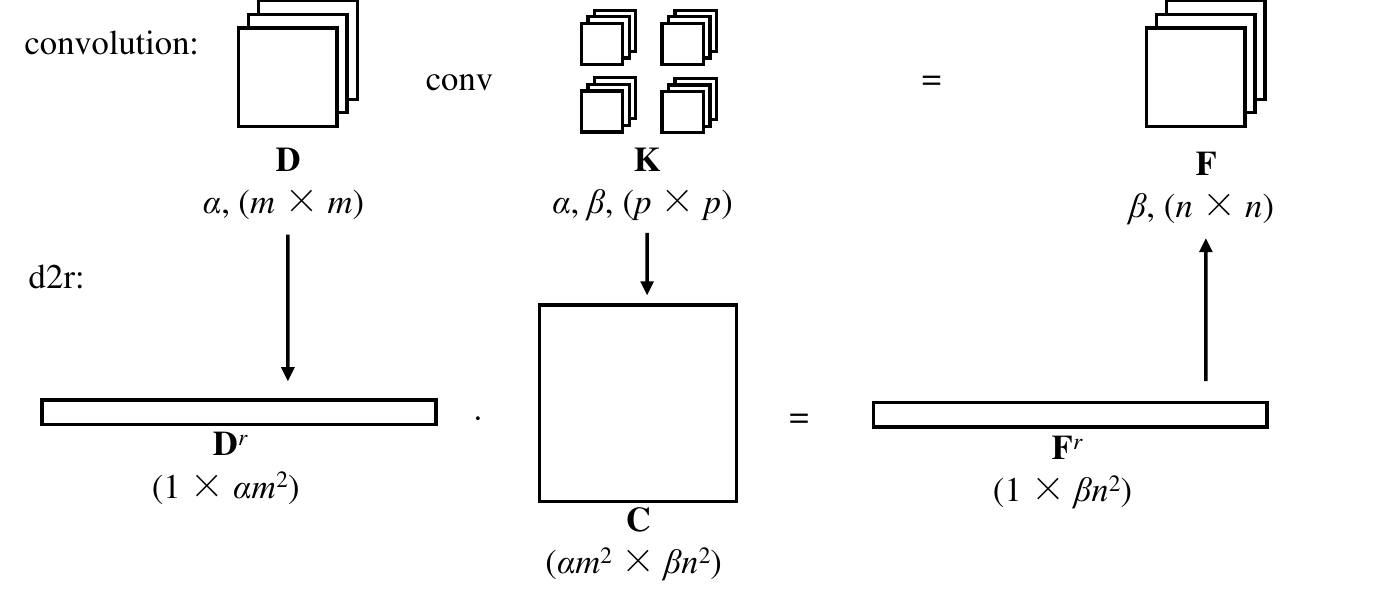}
\caption{Using d2r to compute convolutional layer.}
\label{fig_d2r_full}
\end{figure}

Figure~\ref{fig_d2r_full} shows the change in shapes of matrices when using d2r to compute a convolutional layer.

\begin{figure}[!h]
\centering
\includegraphics{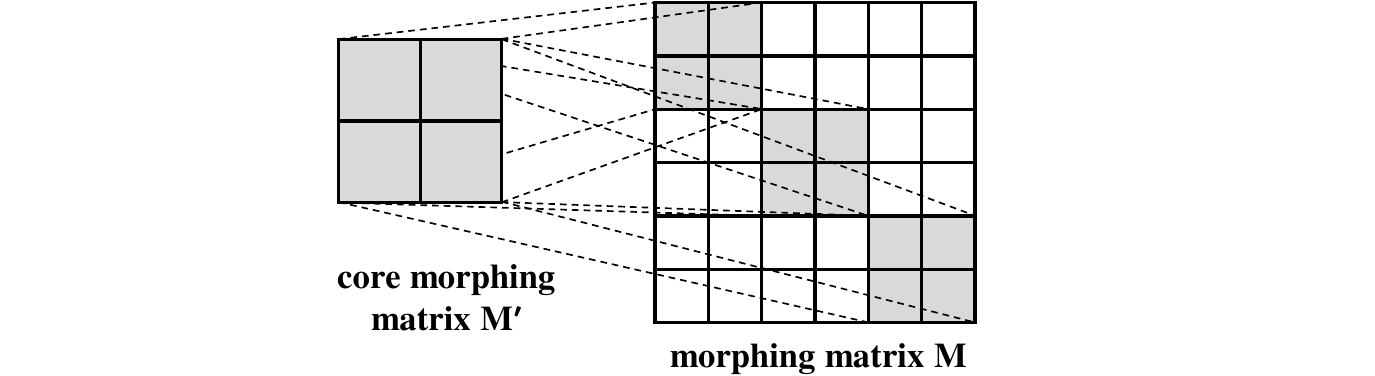}
\caption{An example of diagonally scaling}
\label{fig_diagonally_scaling}
\end{figure}

Figure~\ref{fig_diagonally_scaling} shows an example of diagonally scaling $\textbf{M}'$ with the shape of $2 \times 2$ to compose $\textbf{M}$ with the shape of $6 \times 6$.

\newpage

\begin{figure}[!ht]
\centering
\includegraphics{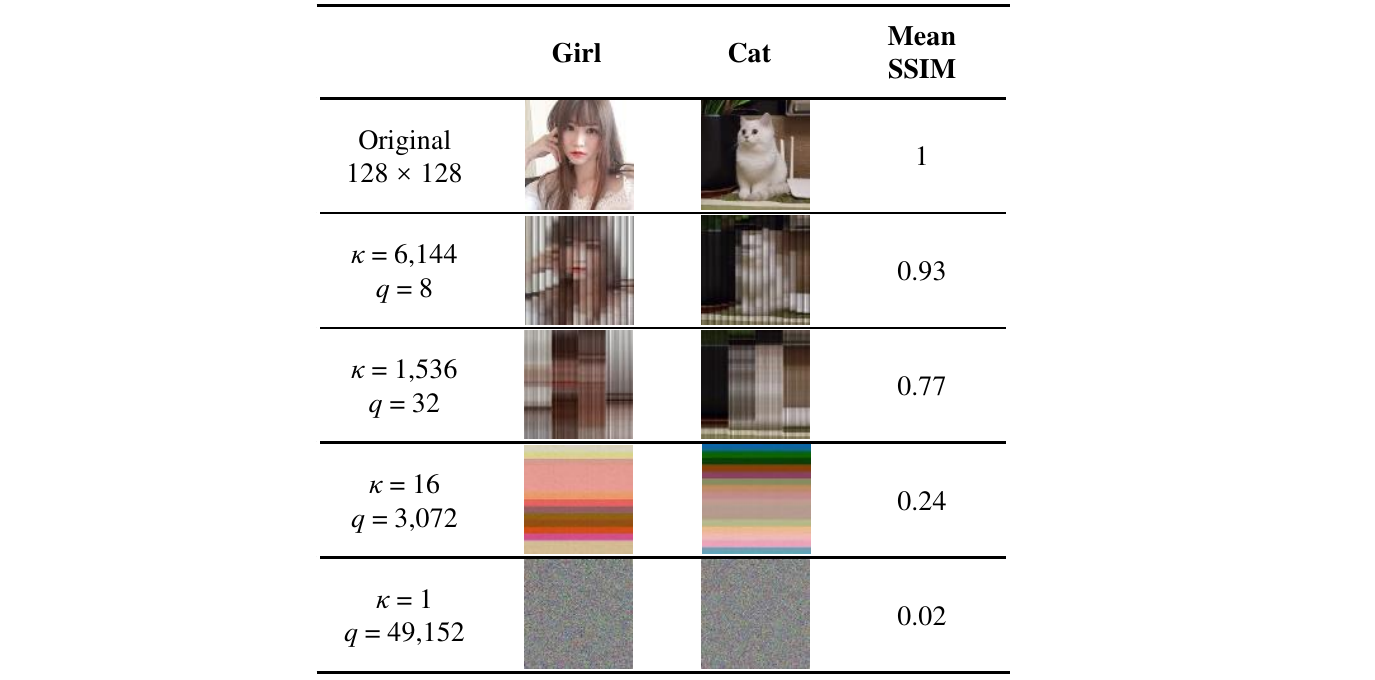}
\caption{$\kappa$ and privacy-preserving effectiveness.}
\label{fig_kappa_and_privacy}
\end{figure}

Figure~\ref{fig_kappa_and_privacy} shows the how change of $\kappa$ affects the privacy-preserving effectiveness of data morphing.
Structural similarity (SSIM) index in the figure is a commonly used method to evaluate the visual similarity between two images.
SSIM index ranges from 0 to 1, and a larger value means higher visual similarity.

\begin{figure}[!ht]
\centering
\includegraphics{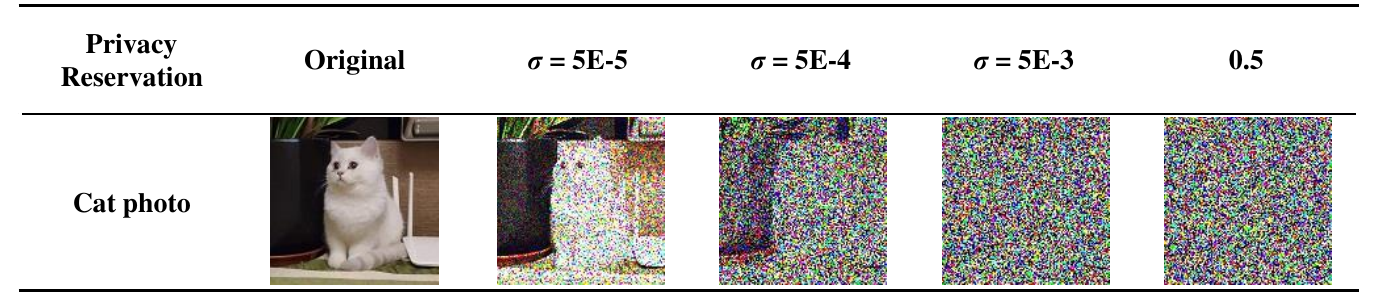}
\caption{Cat photos with different privacy reservation limits.}
\label{fig_cat_photos}
\end{figure}

Figure~\ref{fig_cat_photos} shows the recovered photos with different privacy reservation limits $\sigma$. 
It shows that $\sigma = 0.5$ is a very strict requirement for privacy-preserving.

\end{document}